# Oil Price Trackers Inspired by Immune Memory

William Wilson*, Phil Birkin*, and Uwe Aickelin*

* School of Computer Science, University of Nottingham, UK

wow, pab, uxa@cs.nott.ac.uk

**Abstract**

We outline initial concepts for an immune inspired algorithm to evaluate and predict oil price time series data. The proposed solution evolves a short term pool of trackers dynamically, with each member attempting to map trends and anticipate future price movements. Successful trackers feed into a long term memory pool that can generalise across repeating trend patterns. The resulting sequence of trackers, ordered in time, can be used as a forecasting tool. Examination of the pool of evolving trackers also provides valuable insight into the properties of the crude oil market.

## 1 Introduction

The investigation of time series data to predict future information is a well studied area of research. This paper proposes an immune inspired solution to this problem. Inspiration for memory development was taken from the biological theory proposed by Dr Eric Bell. The theory indicates the existence of two separately identifiable memory populations (Bell, 2005), one long term and the other short term. Their differing characteristics make them ideal in recognising long and short term trends prevalent in time series data. These trends can then be sequenced for use in forecasting and prediction.

## 2 Development of long and short term memory

The flexible learning approach offered by the immune system is attractive as an inspiration for problem solving. However without an adequate memory mechanism the knowledge gained from the learning process would be lost. Memory therefore represents a key contributing factor in the success of the immune system. A difficulty arises in extracting immune memory properties however, because very little is still known about all the biological mechanisms underpinning memory development (Wilson and Garrett, 2004). Theories such as antigen persistence and long lived memory cells (Perelson and Weisbuch, 1997), idiotypic networks, and homeostatic turnover of memory cells (Yates and Callard, 2001) have all attempted to explain the development and maintenance of immune memory but all have been contested.

The attraction of the immune memory theory proposed by Dr Eric Bell is that it provides a simple, clear and logical explanation of memory cell development (Bell, 2005). This theory highlights the evolution of two separate memory pools. The first is a short term memory pool containing short lived, highly proliferative, activated cells that have experienced an antigen. The purpose of this pool is to drive the affinity maturation process to cope with the huge diversity of potential antigen mutations. The second pool consists of those short term memory cells that have evolved to homeostatically turnover to sustain knowledge of an antigen experience over the long term. This long term pool identifies and maintains knowledge of more generalised antigen trends.

## 3 Analysis of oil price trends

The price of WTI crude oil (a world marker price for oil price movements) was selected as the time series for investigation. This data set was chosen because there is considerable economic, financial and government interest in investigating oil price forecasting due to its influence on so many other market sectors. In addition, oil prices have historically exhibited a number of short and long term trend patterns which could map to our long and short term memory concepts, providing an ideal case study for this analysis.

## 4 An immune inspired forecasting solution

The proposed solution comprises a population of "trackers" that correspond to B cells from the immune system. The trackers attempt to identify and record trends in the oil price data. Price data, as measured by the change in price from one time period to the next, is encapsulated within an artificial antigen object and presented to the population of trackers. The antigens are constructed to show current and historical price movements over a particular period. In order to recognise price trends over time, each tracker is allocated a random length "review period". This allows the tracker population to identify a variety of potential price movements over a range of time intervals.

Following the traditional clonal selection approach (de Castro and Von Zuben, 2002), trackers attempt to bind to antigens, and undergo proliferation if successful. The resulting clones mutate in relation to the strength of the bind, with mutation taking one of three forms. One subset of clones has a random price value within their review periods mutated from its original value. A second subset has their review period extended by the addition of a randomly generated price movement to anticipate future potential price movements. A third subset of clones has a random price value removed from their review period to allow them to attempt a better fit to previously experienced antigens.

The degree proliferation is proportional to the strength of the bind and the length of the bound tracker. Initially trackers have relatively short review periods, to enable them to assess a wide variety of price trends. If successful, trackers proliferate and the review periods lengthen to anticipate additional price movements. Excessively long tracker review periods are prevented because trackers become more specific as they lengthen and are therefore less likely to bind. Without successful binds these trackers are likely to be removed via apoptosis. This leads to the evolution of a dynamic population of trackers.

The population of proliferating trackers can be seen to represent the short term memory of experienced price data, as knowledge of an identified price trend is carried forward through the generations of tracker clones. Interrogating the composition of this memory pool provides valuable insight into the dynamics of the oil market.

The process of filtering the short term memory pool to a long term memory subset is achieved through development of the "'tracker sequence'". The tracker sequence is a list of trackers, ordered in time, that best represents the data presented up to the current point in time. Dominant tracker candidates, based on their degree of proliferation, are selected from the short term memory pool and transferred to the tracker sequence for use as a source of long term memory. Generalisations can be made in the tracker sequence for repeating patterns of trackers to highlight recurring price trends. The tracker sequence provides the forecasting mechanism in the system. When new price data becomes available the tracker sequence is examined to identify whether a previously identified trend is recurring again.

## 5 Conclusion

Inspiration was taken from the principles of memory within the immune system to build a system that would identify trends within an oil price time series. This data showed evidence of short term price fluctuations as well as exhibiting underlying long term trends. Detailed inspiration was taken from the theory of immune memory proposed by Dr Eric Bell which identifies two forms of memory, short term and long term. We indicate that these could in principle provide a mechanism to identify and map the short and long term trends evident in the crude oil market which could then be used for forecasting.

## Acknowledgements

The authors would like to thank Dr Eric Bell from the University of Manchester for his valuable input.

## References

E. Bell. University of Manchester, personal communication, 2005.

L. N. de Castro and F. J. Von Zuben. Learning and optimization using the clonal selection principle. *IEEE Transactions on Evolutionary Computation*, 6(3):239–251, 2002.

A. S. Perelson and G. Weisbuch. Immunology for physicists. *Rev. Modern Phys.*, 69:1219–1267, 1997.

W. Wilson and S. Garrett. Modelling immune memory for prediction and computation. In *3rd International Conference in Artificial Immune Systems (ICARIS-2004)*, pages 386–399, Catania, Sicily, Italy, September 2004.

A. Yates and R. Callard. Cell death and the maintenance of immunological memory. *Discrete and Continuous Dynamical Systems*, 1:43–59, 2001.